\documentclass[10pt,twocolumn,letterpaper]{article}

\usepackage{cvpr}              %

\usepackage{float}%
\usepackage{todonotes}
\usepackage{graphicx}
\usepackage{subcaption}
\usepackage{multirow}
\usepackage[percent]{overpic}
\usepackage{xcolor}
\usepackage{listings}
\usepackage[accsupp]{axessibility}
\newcommand{\methodwithspace}{\textsc{Peekaboo} }
\newcommand{\method}{\textsc{Peekaboo}}

\definecolor{cvprblue}{rgb}{0.21,0.49,0.74}
\definecolor{fgorange}{HTML}{f79f79}
\definecolor{bggreen}{HTML}{4b6f4e}
\usepackage[pagebackref,breaklinks,colorlinks,citecolor=cvprblue]{hyperref}
\usepackage{comment}

\def\paperID{7909} %
\def\confName{CVPR}
\def\confYear{2024}

\title{\method: Interactive Video Generation via Masked-Diffusion}

\author{
   Yash Jain${^{1 \dagger}}$
   Anshul Nasery$^{^{2 \dagger}}$\hspace{2mm} Vibhav Vineet$^1$\hspace{2mm} Harkirat Behl$^1$\\[3mm]
  $^1$Microsoft \hspace{3pt}
  $^2$University of Washington \hspace{3pt}\\ %
  \\ [-3mm]
  {\hspace{3pt} 
  \href{https://jinga-lala.github.io/projects/Peekaboo/}{ Project Webpage}}
}

\begin{document}
\newcommand{\TeaserFiveImageGrid}[6]{%
    \centering
    \subfloat{\includegraphics[width=0.2\textwidth]{#2.png}}
    \subfloat{\includegraphics[width=0.2\textwidth]{#1/#3.png}}\hspace{0.2mm}
    \subfloat{\includegraphics[width=0.2\textwidth]{#1/#4.png}}\hspace{0.2mm}
    \subfloat{\includegraphics[width=0.2\textwidth]{#1/#5.png}}\hspace{0.2mm}
    \subfloat{\includegraphics[width=0.2\textwidth]{#1/#6.png}} \newline
}

\twocolumn[{%
\renewcommand\twocolumn[1][]{#1}%
\maketitle
\vspace{-3.5em}
\begin{center}
    \centering
    \includegraphics[width=\textwidth]{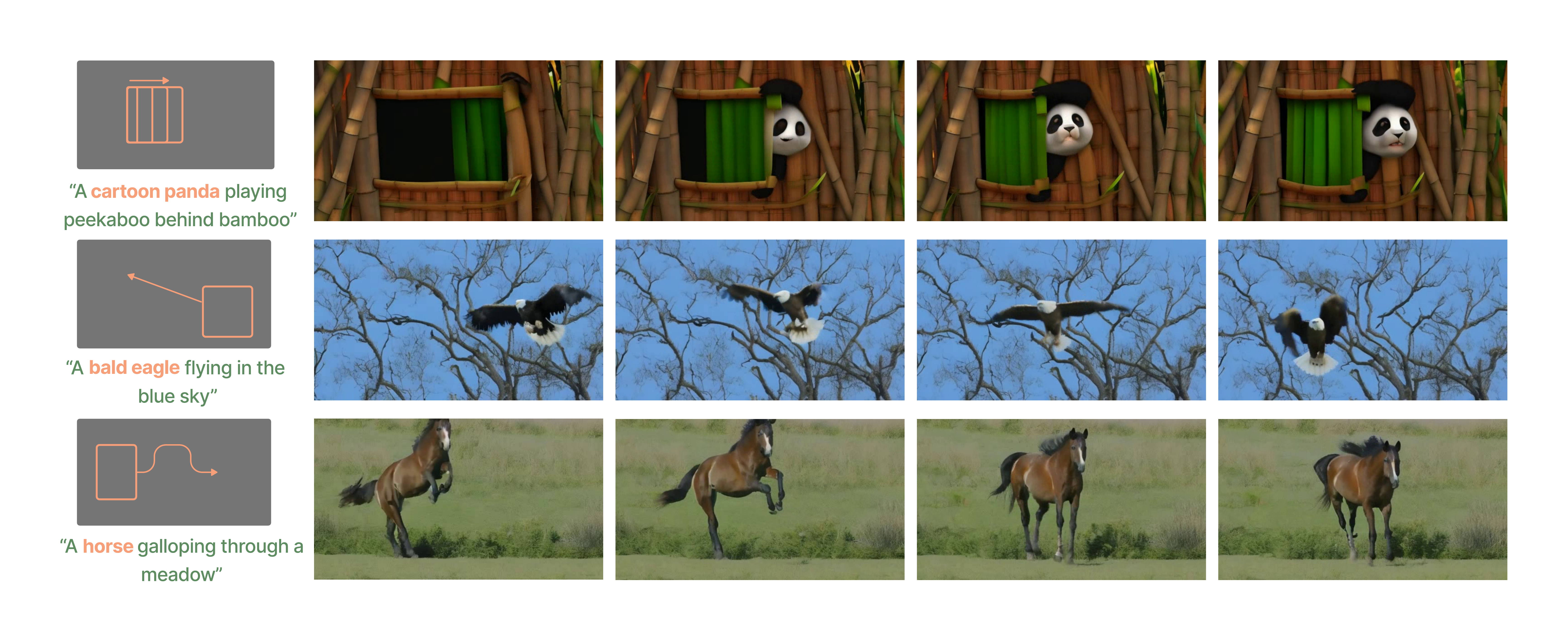}
    
\vspace{-3mm}
\captionof{figure}{\textbf{Zero-training No-latency interactive video generation.} \methodwithspace allows users to control the output (object size, location and trajectory) for any off-the-shelf video diffusion models, through specially designed masking modules. 
First row shows a panda playing \methodwithspace by following an expanding mask in left direction.
}
\label{fig:intro}
\end{center}%
}]

\maketitle
\begin{abstract}
Modern video generation models like Sora have achieved remarkable success in producing high-quality videos.
However, a significant limitation is their inability to offer interactive control to users, a feature that promises to open up unprecedented applications and creativity.
In this work, we introduce the first solution to equip diffusion-based video generation models with spatio-temporal control.
We present \method, a novel masked attention module, which seamlessly integrates with current video generation models offering control without the need for additional training or inference overhead.
To facilitate future research, we also introduce a comprehensive benchmark for interactive video generation. 
This benchmark offers a standardized framework for the community to assess the efficacy of emerging interactive video generation models.
Our extensive qualitative and quantitative assessments reveal that \methodwithspace achieves up to a $3.8\times$ improvement in mIoU over baseline models, all while maintaining the same latency.
Code and benchmark are available on the webpage.      
\end{abstract}

\def\thefootnote{$\dagger$}\footnotetext{equal contribution}
\def\tabQA#1{
\begin{table}[#1]
\setlength{\tabcolsep}{4pt}
\centering
\begin{tabular}{lc}\toprule
Method  & FVD@MSR-VTT ($\downarrow$) \\ \midrule
CogVideo (English) \citep{hong2022cogvideo}  & 1294 \\
MagicVideo \citep{zhou2022magicvideo} & 1290 \\
\midrule
ModelScope 
\citep{wang2023modelscope} & 868 \\
ModelScope w/ \methodwithspace
& 609 \\

\bottomrule
\end{tabular}
\caption{\textbf{Video quality evaluation.} 
\methodwithspace is able to generate videos with higher quality than other baselines. We use bounding boxes generated by GPT-4 as inputs to the model. } 
\label{tab:qa}
\end{table}
}

\def\tabSpatialControl#1{
\begin{table}[#1]
\vspace{-0.1in}
\caption{\textbf{Spatial Control evaluation.} \methodwithspace is able to generate videos with higher spatial control. Moreover, due to spatial control and separate foreground generation \methodwithspace ensures superior meaningful generation (coverage) over the prompt test-set across different models. }
\label{tab:spatial-control}
\setlength{\tabcolsep}{4pt}
\centering
\begin{small}
\begin{tabular}{lccc}\toprule
Method  & \methodwithspace &  mIoU \% ($\uparrow$) & Coverage \% ($\uparrow$)\\ \midrule
Hotshot-XL \cite{Mullan_Hotshot-XL_2023} & - & 17.4 & 69.2 \\
Hotshot-XL & \checkmark & 18.7 & 75.7 \\
Zeroscope \cite{wang2023modelscope}& -  & 12.9 & 37.9 \\
Zeroscope & \checkmark & 31.9 & 47.8 \\
Modelscope \cite{wang2023modelscope} & - & 12.2 & 41.9 \\
Modelscope & \checkmark & 32.6 & 59.9 \\

\bottomrule
\end{tabular}
\end{small}

\end{table}
}

\def\tabInteractive#1{
\begin{table}[#1]
\vspace{-0.1in}
\caption{\textbf{Spatial Control evaluation with custom prompts (stationary).} We generate 25 prompts and a set of 3 bounding boxes for each prompt. We find that \methodwithspace can generate videos with interactive control. %
\label{tab:interactive-control}}
\setlength{\tabcolsep}{4pt}
\centering
\begin{small}
\begin{tabular}{lccc}\toprule
Method  & \methodwithspace &  mIoU \% ($\uparrow$) & Coverage \% ($\uparrow$)\\ \midrule
Zeroscope \cite{wang2023modelscope}& -  & 17.8 & 100\% \\
Zeroscope & \checkmark & 33.2 & 100\% \\

\bottomrule
\end{tabular}
\end{small}

\end{table}
}

\vspace{-3mm}
\section{Introduction}
Generating realistic videos from natural language descriptions is a challenging but exciting task that has recently made significant progress ~\cite{videoworldsimulators2024, singer2022makeavideo, khachatryan2023text2videozero, wu2023tuneavideo, wang2023modelscope}. This is largely due to the development of powerful generative models and latent diffusion models (LDMs \cite{ldm}), which can produce high-quality and diverse videos from text. These models have opened up new possibilities for creative applications and expression.

As the generation quality continues to improve, we can expect more innovation and potential in this domain.
An important aspect is to enable more interactivity and user control over the generated videos (or better \textit{alignment}), by allowing the user to control the spatial and temporal aspects of the video, such as the size, location, pose, and movement of the objects. 
This enables users to express their creativity and imagination through generating videos that match their vision and preferences.
It can also be useful for various applications, such as education, entertainment, advertising, and storytelling, where users can create engaging and personalized video content.

While current models are capable of producing temporally and semantically coherent videos, the user cannot  have spatio-temporal control~\cite{wu2023cvpr}. 
Moreover, these models sometimes fail to produce the main object in the video~\cite{agarwal2023astar}. 
In order to control the output of videos interactively, a model would need to incorporate inputs about spatial layouts into its generation process. 
One set of approaches to achieve spatial control on the network output involves training the entire network or specialized adaptors on spatially grounded data~\cite{mou2023t2i, wang2023videocomposer}. However, such methods involve re-training which is resource and data intensive, limiting their access to the wider community.
This raises the question - Can we create a training-free technique that can introduce interactivity through desired control in videos while utilising large scale pretrained Text-to-Video (T2V) models? 

In this work, we propose \method, a training-free method to augment any off-the-shelf LDM based video-generation model with spatial control. Further, our method has negligible inference overhead. 
For control over individual object generation, we propose to use local context instead of global context.
We propose an efficient strategy to achieve controlled generation within the T2V inference pipeline.
\methodwithspace works by refocusing the spatial-, cross-, and temporal-attention in the UNet~\cite{unet} blocks. 

Figures~\ref{fig:intro}, \ref{fig:static_spatial} and \ref{fig:dynamic_spatial} demonstrate outputs that our method produces for a variety of masks and prompts. Our method is able to maintain a high quality of video generation, while controlling the output spatio-temporally. To evaluate the spatio-temporal control of video generation method, we propose a new benchmark by adapting an existing dataset~\cite{mahdisoltani2018ssv2}, and curating a new dataset for our task (Section~\ref{sec:spatial_control}), and proposing an evaluation strategy for further research in this space. Finally, we show the versatility of our approach on two text-to-video models~\cite{wang2023modelscope} and a text-to-image model~\cite{Rombach_2022_CVPR}. This demonstrates the wide applicability of our method. In summary:
\begin{itemize}
    \item We introduce \methodwithspace which i) allows interactive video generation by inducing spatio-temporal and motion control in the output of any UNet based off-the-shelf video generation model, ii) is \textit{training-free} and iii) adds no additional latency at inference time.
    \item We curate and release a public benchmark, SSv2-ST for evaluating spatio-temporal control in video generation. Further, we create and release the Interactive Motion Control (IMC) dataset to evaluate interactive inputs from a human.
    \item We extensively evaluate \methodwithspace on i) multiple evaluation datasets, ii) with  multiple T2V models (ZeroScope and ModelScope) and iii) multiple evaluation metrics. 
    Our evaluation shows upto $2.9\times$ and $3.8\times$ gain in mIoU score by using \methodwithspace over ZeroScope and ModelScope respectively.
    \item 
    We present qualitative results on spatio-temporally controlled video generation with \method, and 
    also showcase its ability of to overcome some fundamental failure cases present in existing models.
\end{itemize}

\section{Related Work}
\subsection{Video Generation}
Text-based video generation using latent diffusion model has taken a significant leap in recent years \cite{videoworldsimulators2024,singer2022makeavideo, ho2204video, zhou2022magicvideo, ho2022imagen, chen2023controlavideo}. Make-a-video \cite{singer2022makeavideo} introduced the 3D UNet architecture, by decomposing attention layers into spatial, cross and temporal attention layers. Further progress in this generation pipeline was made by \cite{hong2022cogvideo, zhou2022magicvideo, ho2022imagen, chen2023controlavideo}, while keeping the core three attention-layer architecture intact. Although these works focus on generating videos with high relevance to the text input, they do not provide spatio-temporal control in each frame.
More recent works have tried to equip models with this ability to control generation spatio-temporally. Such methods have integrated guidance from depth maps~\cite{esser2023structure}, target motion~\cite{chen2023motionconditioned, hu2023lamd} or a combination of these modalities to generate videos~\cite{wang2023videocomposer}.
However, all these works either require re-training the base model or an external adapter with aligned grounded spatio-temporal data, which is a challenging and expensive task. 

On the other hand, zero-training works include Text2video-zero \cite{khachatryan2023text2videozero}, which integrates optical flow guidance with image model to get consistent frames, ControlVideo \cite{zhang2023controlvideo}, which incorporates sequence of supervising frames (depth maps, stick figures etc.) to control the motion of the video, and  Free-Bloom \cite{huang2023free}, which combines a large language model (LLM) with a text-to-image model to get coherent videos. However, these methods extend specialized image models which were trained on grounded data, and cannot be used with off-the-shelf video-generation models.
The closest method to our work is a concurrent work~\cite{lian2023llm}. The paper uses an LLM to generate bounding box co-ordinates across scenes for an object in the prompt. They use an off-the-shelf video generation model in conjunction with a special guidance module. However, their work has a latency overhead due to extra steps in special guidance module which is absent in our method. 
    
\begin{figure*}[ht!]
    \centering
    \hspace*{-8mm}\includegraphics[width=1.1\textwidth]{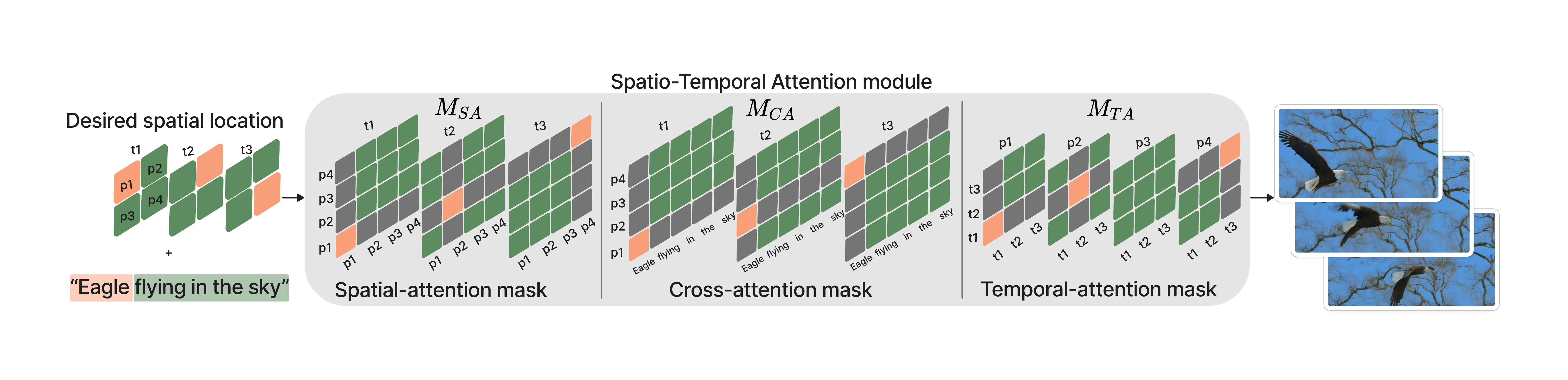}
    \vspace{-12mm}
    \caption{\textbf{\methodwithspace Module}: 
    Our method proposes converting attention modules of an off-the-shelf 3D UNet into masked spatio-temporal mixed attention modules. 
    We propose to use local context for generating individual objects and hence, guide the generation process using attention masks.
    For each of spatial-, cross-, and temporal-attentions, we compute attention masks such that foreground pixels and background pixels attend only within their own region. We illustrate these mask computations for an input mask ( size $2\times2$ and $3$ frames) which changes temporally as shown on the left. \textcolor{bggreen}{Green} pixels are \textcolor{bggreen}{background} pixels and \textcolor{fgorange}{orange} are \textcolor{fgorange}{foreground}. In the attention masks, both \textcolor{bggreen}{green} and \textcolor{fgorange}{orange} pixels have a value of $1$, and \textcolor{gray}{gray} pixels have a value of $0$. We add the colors for ease of exposition. This masking is applied for a fixed number of steps, after which free generation is allowed. Hence, \textcolor{fgorange}{foreground} and \textcolor{bggreen}{background} pixels are hidden from each other before being visible, akin to a game of \method. Best viewed in color. }
    \label{fig:method_figure}
\end{figure*}

\subsection{Controllable Text to Image generation}
Recent works have explored incorporating spatial and stylistic control while generating images from text using diffusion models. These methods can broadly be categorized into those requiring training of the model~\citep{li2023gligen}, and training-free methods~\citep{cao2023masactrl,epstein2023diffusion, phung2023grounded, agarwal2023astar, lian2023llmgrounded}. The former line of works require large amounts of compute resources, as well as spatially grounded data to train their models. The latter either try to shape the spatial and cross-attention maps using energy function guided diffusion, or through masking. Our method is hence closer to the second type of works, however, directly extending these to videos is non-trivial due to the spatio-temporal nature of video generation. 

\paragraph{Guided Attention} The idea of guiding attention maps to control the generation in the image domain has gained popularity recently. 
\citet{agarwal2023astar} focus on minimizing overlap in attention maps for different prompt words, maintaining object information across diffusion steps. \citet{epstein2023diffusion} suggest various energy functions on cross-attention maps to control spatial properties of objects via guided sampling. \citet{phung2023grounded} extend this by ensuring both cross and self-attention maps accurately represent objects, achieving this through optimized noise and segmented attention. Such optimization based methods have inference time overheads, in contrast with our method. \citet{cao2023masactrl} uses thresholded cross-attention maps of the object tokens as masks for self-attention, and ensures that foreground pixels only interact with other pixels within the fore-ground. Their method also requires multiple diffusion inference calls, or requires a source image as an input. Further, they apply their technique only for controlling the pose or actions of objects, which is orthogonal to our task.

\newcommand{\fg}{\text{fg}}
\newcommand{\mbf}[1]{\mathbf{#1}}
\newcommand{\vx}{\mbf{x}}
\newcommand{\vz}{\mbf{z}}
\newcommand{\vc}{\mbf{c}}

\section{Preliminaries: Video Diffusion Models} 
Diffusion models \cite{pmlr-v37-sohl-dickstein15} are generative models that generate images or videos through gradually denoising random gaussian noise. 
The most effective amongst these are Latent Diffusion models (LDMs) \cite{ldm} including Stable Diffusion.
LDMs have two components: First is an image compression auto-encoder, which maps the image $\vx$ to and back from a lower dimensional latent $\vz$.
Second component is a \textit{Denoising Autoencoder} $f_\theta(\vz)$ which operates in the latent space and gradually converts random noise to the image latent. 

\paragraph{Text conditioning} 
Most current text-to-video methods utilize a conditional latent diffusion model which takes a text query as input~\cite{singer2022makeavideo, wang2023modelscope, hotshot}.
The denoising autoencoder is thus conditioned on the text caption $\vc$ as
\begin{equation} \label{eq:denoising}
    \vz_{t+1} = f_\theta (\vz_t | \vc),
\end{equation} 
where $f_\theta$ is a 3D UNet \cite{unet}.
During inference, the input noise is iteratively cleaned and aligned towards the desired text caption.
This is achieved by including a cross attention with the text embedding.

\section{\methodwithspace}
\paragraph{Spatio-temporal conditioning}
For interactive generation, the denoising should also be conditioned on the \textit{user-desired} spatial location and movement of the objects in the video.
This is rather complicated, because unlike Equation \ref{eq:denoising} where the entire latent $\vz_t$ is conditioned on $\vc$, in this setting, parts of the video have to be conditioned on parts of the caption. Note that this would become a conditional distribution with multiple conditions.

A possible solution is to encode the extra conditions as grounding pairs (spatio-temporal volume, text embedding) and pass them as context tokens in the cross attention layer, and train accordingly, taking inspiration from the image based method Gligen \cite{li2023gligen} or even Flamingo \cite{alayrac2022flamingo}. 
On the other hand, we want to explore using a frozen $f_\theta$.

\subsection{Masked Diffusion}
We draw a parallel with the segmentation problem, which is the inverse of spatio-temporal conditioned generation problem. In particular, we take inspiration from MaskFormer \cite{cheng2021perpixel} and Mask2Former \cite{cheng2022maskedattention} who proposed to formulate segmentation as a mask classification problem. This formulation is widely used and accepted, not just for segmentation but even detection \cite{li2022mask} and unified models \cite{zou2022xdecoder}.

\citet{cheng2022maskedattention} propose to split segmentation into grouping into $N$ regions which are represented with binary masks. Hence, \citet{cheng2022maskedattention} advocate using \textit{local} features for segmenting individual objects.
On the other hand, text-to-video diffusion models operate on conditioning a \textit{global} context, as shown in Eqn~\ref{eq:denoising}.
Using the above insight to tackle the problem of spatio-temporal conditioned generation we also propose to use local context for generating individual objects, and then add them together.
In order to control the spatial locations of objects, we propose to modify the attention computations in the transformer blocks of the diffusion model to \textit{masked} attention calls similar to ~\cite{cheng2022maskedattention}. This enables better local generation without any additional computation or diffusion steps.

\subsection{Masked spatio-temporal mixed attention}
Given an input bounding box for a foreground object in the video, we create a binary mask for the foreground object, and downsample it to the size of the latent. We create block sparse attention masks as described below. We use additive masking for attention, i.e. for any query $Q$, key $K$, value $V$ a binary 2D attention mask $M$,
\begin{equation}
\begin{aligned}
    \text{MaskedAttention}&(Q,K,V,M) = \text{softmax}(\frac{QK^T}{d} + \mathcal{M})V \\
\text{where} \quad  
\mathcal{M}[i,j]& = \begin{cases}
                    -\infty \quad \text{if} \ M[i,j] = 0 \\
                    0 \qquad  \text{if} \ M[i,j] = 1
\end{cases}
\end{aligned}
\end{equation}

Here, the additive mask $\mathcal{M}$ is such that it has a large negative value on the masked out entries in $M$, leading to the attention scores for such entries being small. Note that $M \in \{0,1\}^{d_q \times d_k}$, where $d_q, d_k$ are the lengths of queries and keys respectively. We denote the length of the text prompt by $l_{text}$, the length of the video by $l_{video}$, and the dimensions of the latents by $l_{latents}$. The text input is denoted by $T$, and the input mask for frame $f$ is denoted by $M_{input}^f$. For the ease of notation, we assume that the input masks and the latents are flattened along their spatial dimensions. The shape of $M_{input}$ is $l_{video} \times l_{latents}$ We also define the function $\fg(\cdot)$, which takes a pixel or a text token as input, and returns $1$ if it corresponds to the foreground of the video, and $0$ otherwise.

By nudging the foreground token to attend only to the pixels at the desired location at each frame, we can control the position, size and movement of the object. However, naively enforcing this attention constraint only in the cross-attention layer is not sufficient for spatial control. This is because the foreground and background pixels also interact through spatial- and temporal attention. We now discuss how to effectively localise the generation context.

\paragraph{Masked cross attention}
For each frame $f$, we compute an attention mask $M_{CA}^f$, which is a 2-dimensional matrix of size $l_{latents} \times l_{text}$. For each pixel-token pair, this mask is $1$ iff both the pixel and token are foreground, or if both of them are in the background. Formally
\begin{align}
    M_{CA}^f[i,j] &= \fg(M_{input}^f[i]) \ast \fg(T[j]) \nonumber\\
    &+ (1-\fg(M_{input}^f[i])) \ast (1 - \fg(T[j]))
\end{align}
This ensures that the latents attend to the foreground and the background tokens at the correct locations.

\paragraph{Masked spatial attention}
For each frame $f$, we compute an attention mask $M_{SA}^f$ which is a 2-dimensional matrix of size $l_{latents} \times l_{latents}$. For each pixel pair, this mask is $1$ iff both the pixels are foreground, or if both of them are in the background. Formally
\begin{align}
    M_{SA}^f[i,j] &= \fg(M_{input}^f[i]) \ast \fg(M_{input}^f[j]) \nonumber\\
    &+ (1-\fg(M_{input}^f[i])) \ast (1 - \fg(M_{input}^f[j]))
\end{align}
This additionally focuses the attention to ensure that the foreground and background are generated at the correct locations, by encouraging them to evolve independently for the initial steps. This also helps improve the quality of generation since it leads to adequate interaction within the foreground and background regions. A similar idea in the context of image generation had been explored in MasaCtrl\cite{cao2023masactrl} in their self attention layer.

\paragraph{Masked temporal attention}
For each latent pixel $i$, we compute a mask $M_{TA}^i$, which is a 2D matrix of size $l_{video} \times l_{video}$. For each frame pair, the value of this mask is $1$ if the pixel $i$ is a foreground pixel in both frames, or if it is a background pixel in both frames. Formally,
\begin{align}
    M_{TA}^i[f,k] &= \fg(M_{input}^f[i]) 
    \ast \fg(M_{input}^k[i]) \nonumber\\
    &+ (1-\fg(M_{input}^f[i])) \ast (1 - \fg(M_{input}^k[i]))
\end{align}

This ensures temporal consistency for the generation since it provides correct local context for foreground and background latents across time.

\subsection{Zero-training Pipeline} %

Putting the selective masks in a diffusion pipeline gives us a zero-training method, \textit{dubbed} \method. \methodwithspace integrates in the attention layers of the 3D-UNet architecture of text-to-video models. We perform selective generation of foreground and background object for a fixed number of steps $t$ and then allow free-generation for the rest of steps. This free generation enables the foreground and background pixels to cohesively integrate with each other on the same canvas as have been done by \cite{lian2023llmgrounded, bar2023multidiffusion}. In essence, our method ensures that foreground pixels cannot ``see" the background pixels for some steps (and vice versa), before being visible to each other. This is akin to a game of \method.

Unlike image control methods \cite{lian2023llmgrounded, bar2023multidiffusion}, \methodwithspace does not require extra inference overhead in the form of more number of diffusion steps and works with very low value of fixed step $t$ (refer to Appendix for more details). 
This ensures that there is no gain in latency during generation while providing extensive spatial control. 

Further, since \methodwithspace is a zero-training off-the-shelf technique it is versatile to implement in all diffusion models and can work with present as well as future text-to-video models. Thus, \methodwithspace can give spatio-temporal control in better quality generation models which are not explicitly trained on any spatially-grounded dataset.

\subsection{Extensions}
Currently, majority of the diffusion pipelines have a UNet-based architecture. This enables \methodwithspace to become versatile and be used not only in Text-to-Video scenario, but in Text-to-Image setup with a possibility in other generation modalities too.

\noindent \textbf{Automatically generated input masks}
Since our method is orthogonal to the choice of input masks, we can use a large language model to generate the input masks for an object corresponding to a given prompt, in a similar fashion as concurrent works~\cite{lin2023videodirectorgpt, lian2023llmgrounded}. In Table~\ref{tab:qa}, we demonstrate that doing this leads to videos with better quality than the baseline model. Moreover, it enables our method to be end-to-end in terms of only requiring a text prompt from the user.

\begin{figure}[t!]

    \centering
    \includegraphics[width=\linewidth]{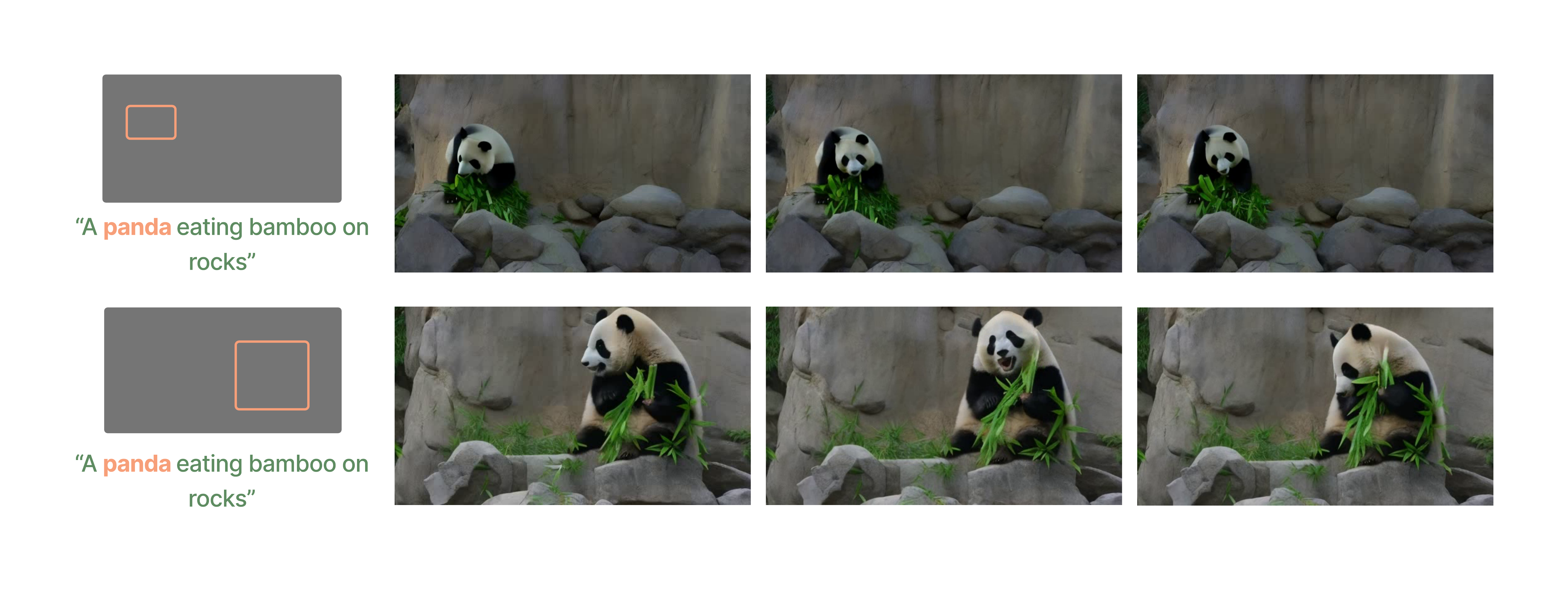}
    \vspace*{-8mm}
    \caption{\textbf{Spatial control with \method:} Changing the bounding box while providing the same prompt leads to generated panda being faithful to the input layout in terms of size and location with our method. } %
    \label{fig:static_spatial}
\end{figure}
\noindent \textbf{Image generation}
Image generation diffusion method are based of 2D-UNet architecture, with the absence of temporal attention layer. Analogous to our text-to-video setup, we can adapt \methodwithspace for Image Diffusion models. The spatial-attention mask maintains the semantic structure of the image while the cross-attention mask focuses the attention of foreground token on desired location and vice versa for background.
In Figure~\ref{fig:text2image}, we showcase spatial control on an off-the-shelf diffusion model and highlight the versatility of our method.

\newcommand{\imagegrid}[6]{%
    \begin{figure*}[ht]
    \centering
    \subfloat{\includegraphics[width=0.23\textwidth]{#1/frame_#2.png}}\hspace{2mm}
    \subfloat{\includegraphics[width=0.23\textwidth]{#1/frame_#3.png}}\hspace{2mm}
    \subfloat{\includegraphics[width=0.23\textwidth]{#1/frame_#4.png}}\hspace{2mm}
    \subfloat{\includegraphics[width=0.23\textwidth]{#1/frame_#5.png}}
    \caption{#6}
    \end{figure*}
}

\newcommand{\FiveImageGrid}[6]{%
    \centering
    \subfloat{\includegraphics[width=0.19\textwidth]{#1.png}}\hspace{0.2mm}
    \subfloat{\includegraphics[width=0.19\textwidth]{#2/#3.png}}\hspace{0.2mm}
    \subfloat{\includegraphics[width=0.19\textwidth]{#2/#4.png}}\hspace{0.2mm}
    \subfloat{\includegraphics[width=0.19\textwidth]{#2/#5.png}}\hspace{0.2mm}
    \subfloat{\includegraphics[width=0.19\textwidth]{#2/#6.png}} \newline
}

\section{Experiments}
In this section, we demonstrate the effectiveness of our method. The main focus of our technique is to generate objects in specific spatio-temporal locations in videos. We first evaluate this region level control in Sec \ref{sec:spatial_control}. 
In \ref{sec:video_quality}, we compare the generation quality against baselines to show that grounding enables much better generation. We also demonstrate qualitative results of our method, and perform ablation analysis on our method and show the effect of each component on the final generations.

\subsection{Quantiative Analysis}
We first present quantitative results on evaluating the spatial control and the quality of videos generated by \method.
\subsubsection{Spatial Control} \label{sec:spatial_control}

\begin{table*}[ht!]
\setlength{\tabcolsep}{1.8pt}
\centering

\begin{footnotesize}
\begin{tabular}[width=\textwidth]{lccccccccccccccccccc}
\toprule 
\multirow{2}{*}{Method} & \multirow{2}{*}{Latency} & \multicolumn{4}{c}{DAVIS16} & \multicolumn{4}{c}{LaSOT} & \multicolumn{4}{c}{ssv2-ST} & \multicolumn{4}{c}{IMC} \\ 
\cmidrule(lr){3-6} \cmidrule(lr){7-10} \cmidrule(lr){11-14} \cmidrule(lr){15-18}
                        &              & mIoU ($\uparrow$)  & AP50 ($\uparrow$)  & Cov ($\uparrow$) & CD ($\downarrow$)
                        & mIoU & AP50 & Cov & CD ($\downarrow$)
                        & mIoU & AP50 & Cov & CD ($\downarrow$)
                        & mIoU & AP50  & Cov & CD ($\downarrow$)\\ 
\midrule
LLM-VD \cite{lian2023llm}*      & 2.20$\times$ 
                                & \textbf{26.1} & 15.2 & 96   & 0.19
                                & 13.5 & 4.6  & 98   & 0.24
                                & 27.2 & 21.2 & 61  & 0.12 
                                & 36.1 & 33.3 & \bf{97}  & 0.13 \\
\midrule 
ModelScope \cite{wang2023modelscope}
                                & 1$\times$               
                                & 19.6 & 5.7  & 100  & 0.25
                                & 4.0  & 0.7  & 96   & \bf{0.33} 
                                & 12.0 & 6.6  & 44.7 & 0.17 
                                & 9.6  & 2.4  & 93.3 & 0.25\\
w/ \method                      & 1.03$\times$       
                                & 26.0 & 16.6 & 93   & \bf{0.18} 
                                & \bf{14.6} & 10.2 & 98   & 0.25 
                                & 33.2 & 35.8 & \bf{63.7} & \bf{0.10} 
                                & 36.1 & 33.3 & 96.6 & 0.13 \\
\midrule ZeroScope \cite{wang2023modelscope}
                                 & 1$\times$               
                                 & 11.7 & 0.1  & 100  & 0.22
                                 & 3.6  & 0.4  & 100  & 0.3 
                                 & 13.9 & 9.3  & 42.0 & 0.22
                                 & 12.6 & 0.6  & 88.0 & 0.26\\
w/ \method                      & 1.03$\times$          
                                & 20.6 & \bf{17.9} & \bf{100}  & 0.19
                                & 11.5  & \bf{11.9}  & \bf{100}  & 0.28  
                                & \bf{34.7} & \bf{39.8} & 56.3 & 0.17 
                                & \bf{36.3} & \bf{33.8} & 96.3 & \bf{0.12}\\
\bottomrule
\end{tabular}%
\end{footnotesize}

\caption{\textbf{Evaluation of spatio-temporal control on mIoU, AP50, Coverage and Centroid Distance (CD):} We evaluate two different video generation models on spatio-temporal control against DAVIS16, LaSOT, ssv2-ST, and IMC datasets. As demonstrated by mIoU and CD, the videos generated by \methodwithspace endow the baselines with spatio-temporal control. \methodwithspace also increases the quality of the main objects in the scene, as seen by higher AP50 and Coverage scores. Further, LLM-VD\cite{lian2023llm} has higher inference cost whereas \methodwithspace does not affect latency. *: LLM-VD\cite{lian2023llm} has not released code, this is our re-implementation.}
\label{tab:spatial-control}
\end{table*}

\noindent \textbf{Evaluation Datasets}
Evaluating spatial control in multiple text-to-video models is a challenging task and requires creating a common benchmark for (prompt, mask) pairs. We develop a benchmark obtained from a public video dataset with high-quality masks that represent realistic locations for day-to-day subjects. Further, we also curated a set of (prompt, mask) pairs that represent an interactive input from the user in controlling a video and its subject.
\begin{itemize}
    \item \textbf{Something-something v2-Spatio-Temporal (ssv2-ST):} We use Something-Something v2 dataset \cite{goyal2017something, mahdisoltani2018ssv2} to obtain the generation prompts and ground truth masks from real action videos. We filter out a set of 295 prompts. The details for this filtering are in the appendix. We then use an off-the-shelf \textit{OWL-ViT-large} open-vocabulary object detector \cite{owlvit} to obtain the bounding box (bbox) annotations of the object in the videos. This set represents bbox and prompt pairs of real-world videos, serving as a test bed for both the quality and control of methods for generating realistic videos with spatio-temporal control. 
    \item \textbf{Interactive Motion Control (IMC):} We also curate a set of prompts and bounding boxes which are manually defined. We use GPT-4 to generate prompts and pick a set of 34 prompts of objects in their natural contexts. These prompts are varied in the type of object, size of the object and the type of motion exhibited. We then annotate 3 sets of bboxes for each prompt, where the location, path taken, speed and size are varied. This set of 102 prompt-bbox pairs serve as our custom evaluation set for spatial control. Note that since ssv2-ST dataset has a lot of inanimate objects, we bias this dataset to contain more living objects. This dataset represents possible input pairs that real users may generate.
    \item \textbf{LaSOT}: We repurpose a large-scale object tracking dataset --LaSOT\cite{fan2020lasot}-- for evaluating control in video generation. This dataset contains prompt-bbox-video triplets for a large number of classes. The videos have frame level annotations specifying the location of the object in the video. We subsample the videos to 8 FPS and then randomly pick up 2 clips per video from the test set of this dataset. This gives us 450 total clips across 70 different object categories.
    \item \textbf{DAVIS-16}: DAVIS-16\cite{Perazzi2016davis} is another video object segmentation dataset that we consider. We take videos from its test set, manually annotating them with prompts. We use the provided segmentation masks to create input bboxes. This gives us 40 prompt-bbox pairs in total, where each video has a different subject. 
\end{itemize}
\tabQA{t!}
\paragraph{Experimental Setup} We use two base models for our evaluation, Zeroscope and ModelScope~\cite{wang2023modelscope}. 
These models are run for the default number of inference steps, with default temperature and classifier guidance parameters. We also experiment with mask guidance steps in the appendix. We provide the model with the text prompt and the set of input bboxes. The generated videos are then evaluated for spatio-temporal control and video quality.

\noindent \textbf{Evaluation methodology.} After generating videos for each (prompt, mask) pair, we pass these videos through \textit{OWL-ViT-large} detector to compute bboxes for each generated video. 
We first compute the fraction of generated videos for which OwL-ViT detects bboxes in more that 50\% of the generated frames. We report this fraction as the \textit{Coverage} of the model in Table~\ref{tab:spatial-control}. However, the lack of a detected bbox does not necessarily imply the lack of an object generated, since OwL-ViT could fail to capture some objects correctly. Hence, to evaluate the spatio-temporal control of the generation method, we first filter out videos where less than 50\% frames have a detected bbox. We then compute the Intersection-over-Union of the detected bboxes and the input mask on these  filtered videos. We report the mean of these IoU (\textit{mIoU}) scores for each method in Table~\ref{tab:spatial-control}. These two metrics together provide a good proxy of the quality of the generated videos as well as the spatio-temporal control imparted. We compute the \textit{Centroid Distance (CD)} as the distance between the centroid of the generated object and input mask, normalized to 1. This measures control of the generation location. Finally, we report the average precision@50\% (\textit{AP50}) of the detected and input bboxes averaged over all videos. For generated frames with the object present, AP50 represents the spatial control provided by the method, while mIoU measures the model's ability to match the input bboxes exactly and penalizes frames where the object cannot be detected.

\begin{figure}[t]
    \hfill %
    \begin{subfigure}[b]{0.48\linewidth} %
        \includegraphics[width=\linewidth]{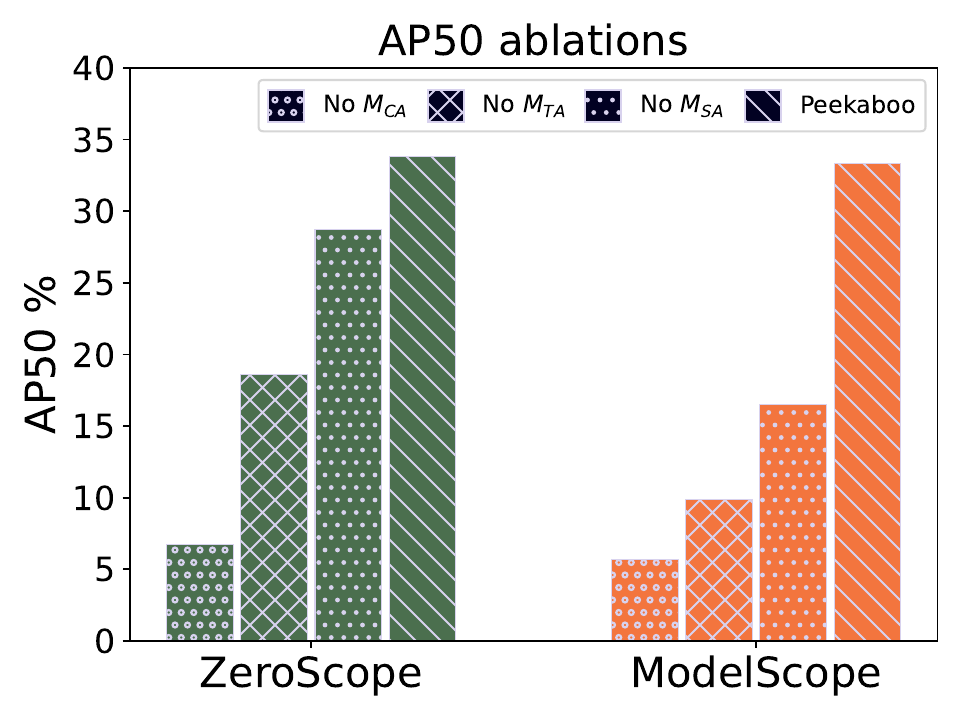} %
    \end{subfigure}
    \hfill
    \begin{subfigure}[b]{0.48\linewidth} %
        \includegraphics[width=\linewidth]{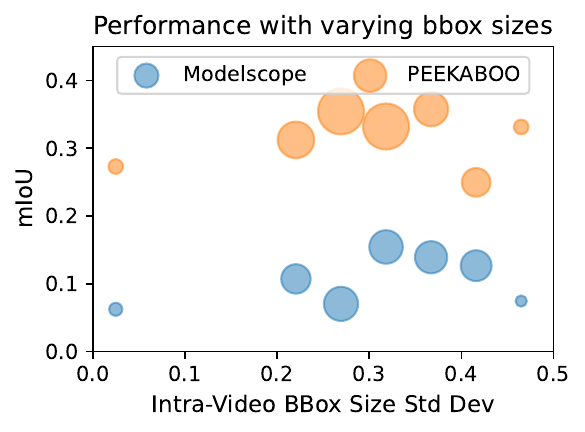} %
    \end{subfigure}
    \caption{(a) \textbf{Ablation Studies on IMC}: The performance of \methodwithspace varies as different attention masks are removed. The AP50 drops the most when cross-attention masks are removed, indicating their importance to spatial control, followed by temporal and spatial attention. (b) \textbf{Performance against varying bbox sizes:} mIoU score compared against varying bounding box sizes across time in videos. We observe that \methodwithspace provides better control independent of bbox size variation.  Best viewed in color.}
    \label{fig:ablations}
\end{figure}

\begin{figure*}[t!]
    \centering
    \includegraphics[width=\linewidth]{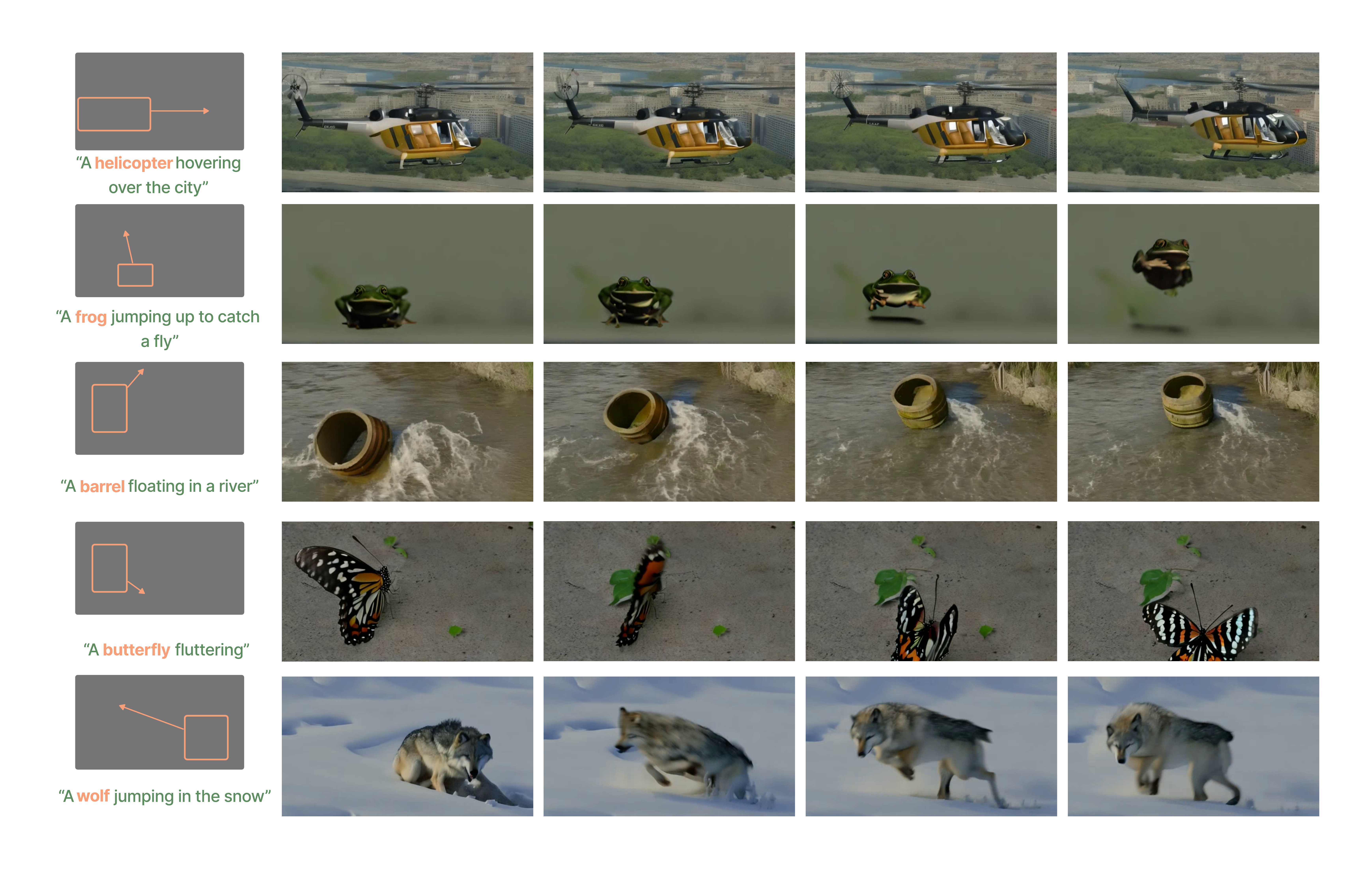}
    \vspace*{-10mm}
    \caption{\textbf{\methodwithspace with a moving mask:} As demonstrated, our method can mimic the input mask trajectories to generated spatio-temporally controlled videos with realistic motions. For \eg in the last row, the wolf is jumping following the mask on the left.} %
    \label{fig:dynamic_spatial}
\end{figure*}

\noindent \textbf{Results} In Table~\ref{tab:spatial-control}, we demonstrate that our method adds control to the model. We verify that our method enables spatio-temporal control, as evidenced by the higher (upto 2.5x) AP@50 scores and lower CF on all four datasets (DAVIS16, LaSOT, ssv2-ST and IMC). This means that the generated objects are close to the true centroid of the input mask, and their shape and size are also consistent with the input mask. We observe significant jump in mIoU score with \methodwithspace across different models and LLM-VD \cite{lian2023llm}, highlighting superior spatio-temporal control achieved through \method. Further, \methodwithspace has a higher coverage than the baseline models and LLM-VD, indicating that our method is also able to generate objects when the base model could not do so.  Finally, we note that \methodwithspace introduce minimal latency increase to the original method compared to LLM-VD's 2.20$\times$ increase in inference time on ModelScope.

\subsubsection{Quality control} \label{sec:video_quality}

While the above datasets provide evidence for \method's spatio-temporal control, we also benchmark our method on MSR-VTT\cite{xu2016msr-vtt} -- a large scale video generation dataset -- to evaluate the quality of videos generated.  We benchmark \methodwithspace for evaluating quality control using Fr\'echet Video Distance score (FVD) metric \cite{fvd}. FVD is calculated based on I3D model trained on Kinetics-400 dataset \cite{i3dAndKinetics}. Following previous works, we evaluate on the test-set of MSR-VTT  containing 2900 videos by randomly sampling one of the 20 captions for each video.
We demonstrate the versatility of our method by using bounding boxes generated by GPT-4. We query GPT-4 to generate series of locations for the foreground object depending on the prompt. We evaluate on ModelScope model and compare the scores with \method.
Table~\ref{tab:qa} shows that \methodwithspace increases the quality of generated while providing spatial control during video generation.
The performance of these methods is also better than other baselines, indicating that \methodwithspace can be integrated in an automated pipeline to use GPT-4 generated bboxes and output a coherent video.

\begin{figure}[t!]
   \centering
    \hspace{0.1mm}
   \begin{subfigure}{.45\textwidth}
       \centering
       \subfloat{\begin{overpic}[width=0.48\linewidth]{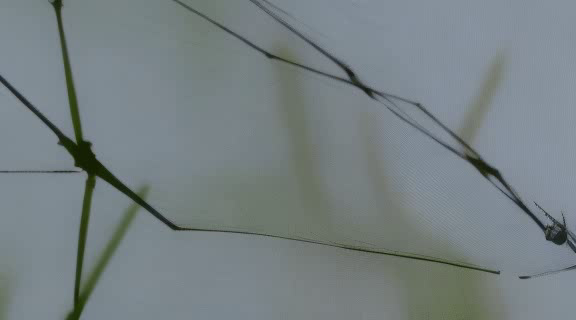}
          \put(67,0){\includegraphics[width=0.15\linewidth]{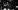}}
       \end{overpic}}
       \hspace{0.1mm}
       \subfloat{\begin{overpic}[width=0.48\linewidth]{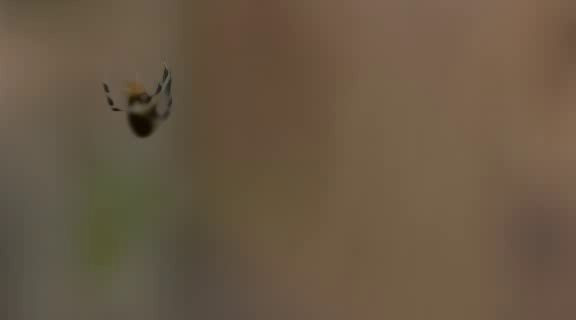}
          \put(67,0){\includegraphics[width=0.15\linewidth]{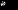}}
       \end{overpic}}
       \caption*{(a) A \textcolor{fgorange}{spider} \color{black} descending on its web. }
   \end{subfigure}
   \newline
   \vspace{1em} %
   \begin{subfigure}{.45\textwidth}
       \centering
       \subfloat{\includegraphics[width=0.48\linewidth]{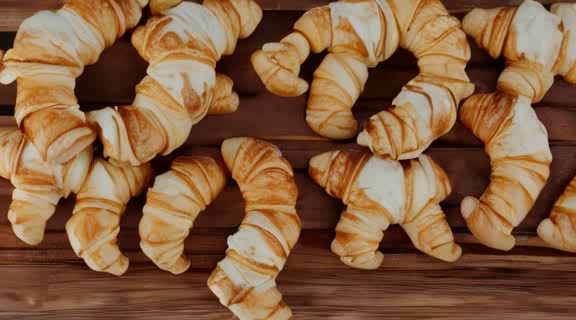}}
       \hspace{0.1mm}
       \subfloat{\includegraphics[width=0.48\linewidth]{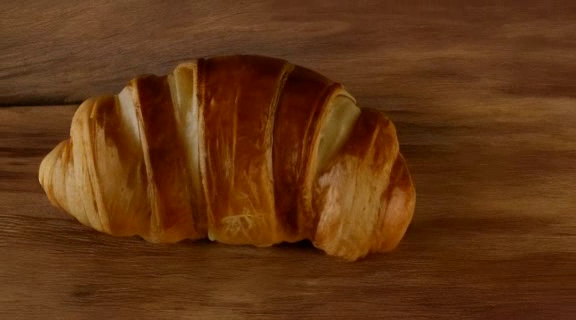}}
       \caption*{(b) A \textcolor{fgorange}{croissant} on a wooden table. }
    \end{subfigure}
    \vspace*{-4mm}
    \caption{\textbf{Overcoming model failures}: Frames on the left are generated by zero-scope, and frames on the right are generated by \method. Inset in the first row are cross-attention map between the word ``spider" and the pixels in the video frame. We can generate objects that are otherwise omitted from the video by the base model. The attention maps also show that explicit masking leads to better generation. The second row depicts a numeracy failure of the baseline where \methodwithspace can control the number of objects. }
    \label{fig:miracle}
\end{figure}

\subsubsection{Ablation analysis}
\noindent \textbf{Ablation on individual attention masks.} A Spatio-Temporal attention block consists of three types of attention layers-- Spatial, Cross and Temporal. \methodwithspace applies masking on all three layers, however, the effect of each mask on the generation quality is different. In this section, we experiment with \methodwithspace by disabling masking for each attention layer one-by-one. We evaluate the AP50 score for ModelScope and ZeroScope on the IMC dataset, as shown in Figure~\ref{fig:ablations}. The performance drops massively when any one of the attention mask is not provided. We observe that not passing $M_{CA}$ hurts the control the most. This is explained by the fact that main object's text token will not focus its attention at the bbox location, leading to the object being generated at a different location. Surprisingly, not passing $M_{TA}$ is worse than not passing $M_{SA}$. We conjecture that removing spatial attention mask leads to degraded videos, while removing the temporal attention mask leads to the loss of temporal control. Since the latter model still generates higher quality objects at incorrect locations, it has a lower AP50 score. We notice that the Coverage of the model after removing $M_{SA}$ is much less than the Coverage of the model after removing $M_{TA}$, providing evidence in support of our hypothesis.

\noindent \textbf{Performance on varying Bbox sizes.} \methodwithspace can accommodate varying bbox sizes within videos. Fig~\ref{fig:static_spatial} shows qualitative control on input bboxes with the output. Moreover, SSv2-ST dataset contain examples with varying bbox sizes. Thus, we further quantify our SSv2-ST results in Fig~\ref{fig:ablations}(b) by plotting mIoU scores against varying box sizes across time in videos. For each prompt of SSv2-ST, we compute the standard deviation of the bbox size (relative to the mean input) to highlight the variation across time. We observe a consistent improvement by \methodwithspace on base model independent of box size variation.

\subsection{Qualitative Results}
In Figure~\ref{fig:intro}, we present examples of videos generated by \methodwithspace applied on ZeroScope~\cite{wang2023modelscope}. As demonstrated, the videos follow the bbox input. Through these qualitative results, we highlight the versatility of bbox input in capturing the shape, size, location and motion, and show how our method can utilize this information interactively.

\noindent \textbf{Static spatial control.} Figure~\ref{fig:static_spatial} shows videos where the object is statically located in the frame. Our method can control the position of the object, and can also change the size of the object as specified by the user through a bbox.

\noindent \textbf{Dynamic spatial control.} Figure~\ref{fig:dynamic_spatial} present videos where the main subject is moving on a desired path. Our method generated realistic looking movements for various motion trajectories. The temporal masking of our method also enables it to handle cases where the mask disappears mid-way through the scene, as is the case in the first row in Figure~\ref{fig:intro}, while the spatial and cross-attention masking ensures spatial coherence of the generated frames with the input bounding boxes.

\noindent \textbf{Overcoming model failures.}
Diffusion models can have a bias on their generation capabilities depending on their training data. However, we observe that \methodwithspace can suppress those biases and produce high quality generation by forcing the model to generate foreground object at a specific location. In Figure~\ref{fig:miracle}, we present results of prompts where the original model fails to produce the foreground object however, our method can produce the object in the user specified location and motion. The inset figures in Figure~\ref{fig:miracle} reveal the reason for this -- while the cross-attention corresponding to the word ``spider" is diffused across the entire canvas in the original model, \methodwithspace focuses this attention on the desired region. Further, Figure~\ref{fig:miracle} depicts the example of hallucination by generation model where the subject was generated multiple times. Again, \methodwithspace solves this issue due to spatial-attention mask and cross-attention at a specific location.

\section{Conclusion}
In this work, we explore interactive video generation. 
We hope that this work will inspire more research in this area.
To this end, we propose a new benchmark for this task and \method, which is a training-free, no latency overhead method to endow video models with spatio-temporal control.
Future work involves exploring \methodwithspace for image-to-video generation, video-to-video generation and long form video generation.

{
    \small
    \bibliographystyle{ieeenat_fullname}
    \bibliography{main}
}
\newpage
\clearpage
\appendix

\section{Text to image synthesis}
\begin{figure}[t!]
   \centering
    \begin{subfigure}{.23\textwidth}
       \centering
    \subfloat{\begin{overpic}[width=0.48\textwidth]{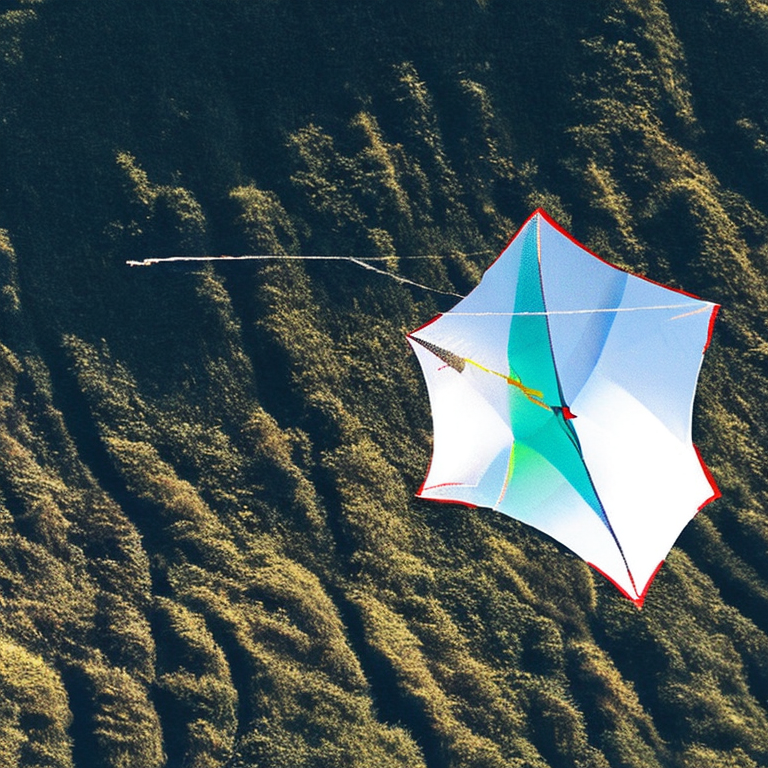}
      \put(0,0){\includegraphics[width=0.1\textwidth]{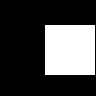}}
    \end{overpic}} \hspace{0.1mm}
    \subfloat{\begin{overpic}[width=0.48\textwidth]{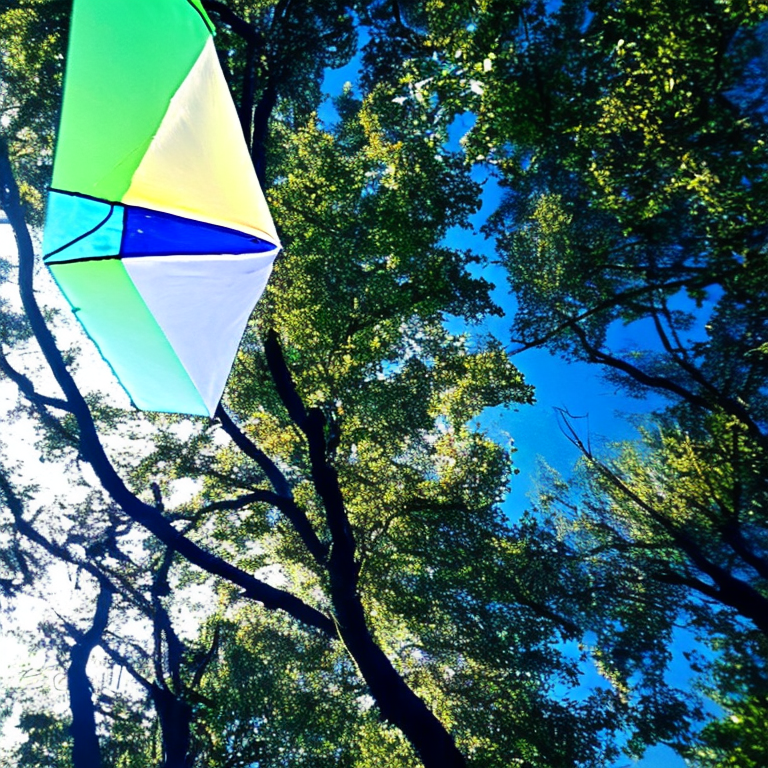}
      \put(0,0){\includegraphics[width=0.1\textwidth]{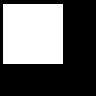}}
    \end{overpic}} 
   \caption*{(a) A \textcolor{fgorange}{kite} flying in the sky. }
   \end{subfigure}    
    \hspace{-0.5mm}
       \begin{subfigure}{.23\textwidth}
       \centering
    \subfloat{\begin{overpic}[width=0.48\textwidth]{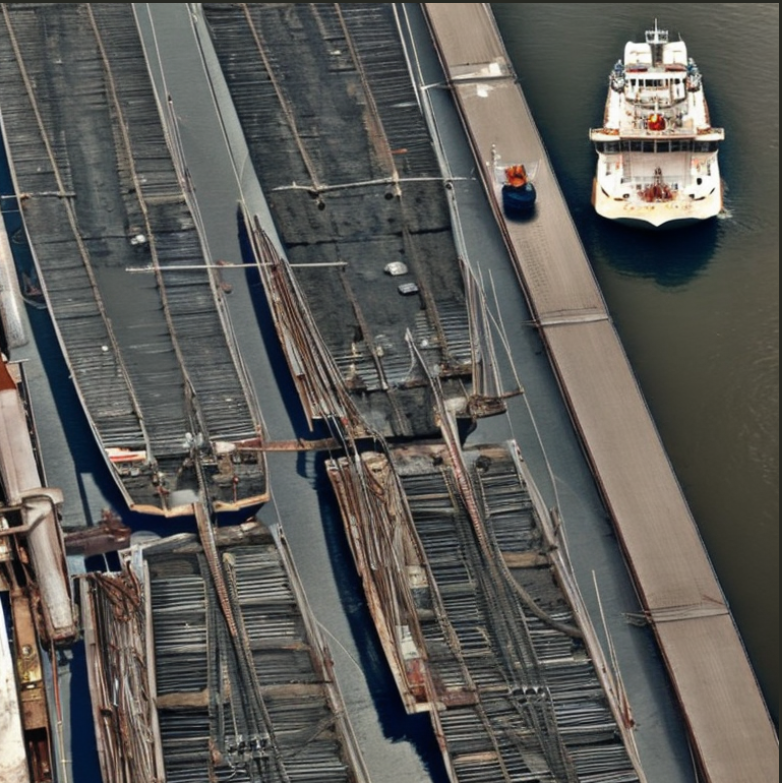}
      \put(0,0){\includegraphics[width=0.1\textwidth]{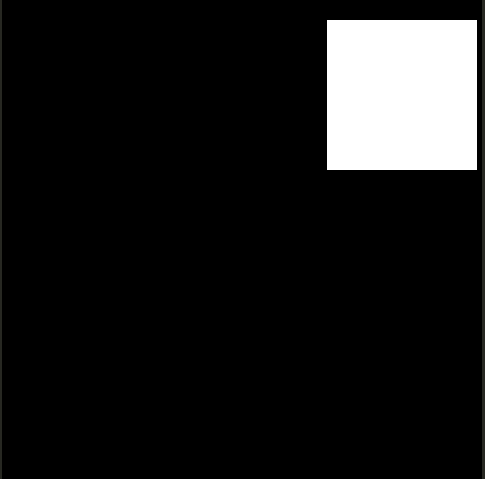}}
    \end{overpic}} \hspace{0.1mm}
    \subfloat{\begin{overpic}[width=0.48\textwidth]{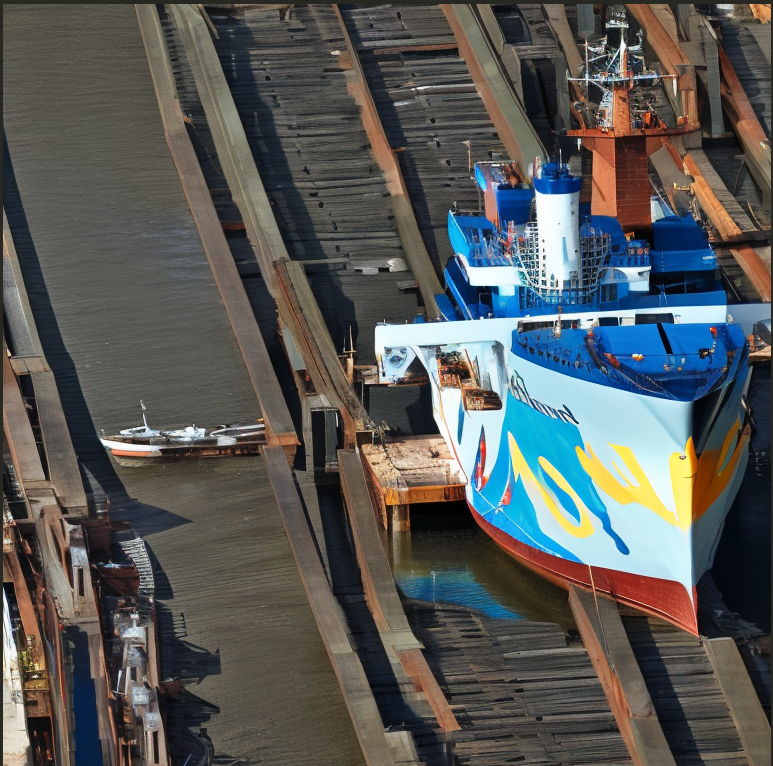}
      \put(0,0){\includegraphics[width=0.1\textwidth]{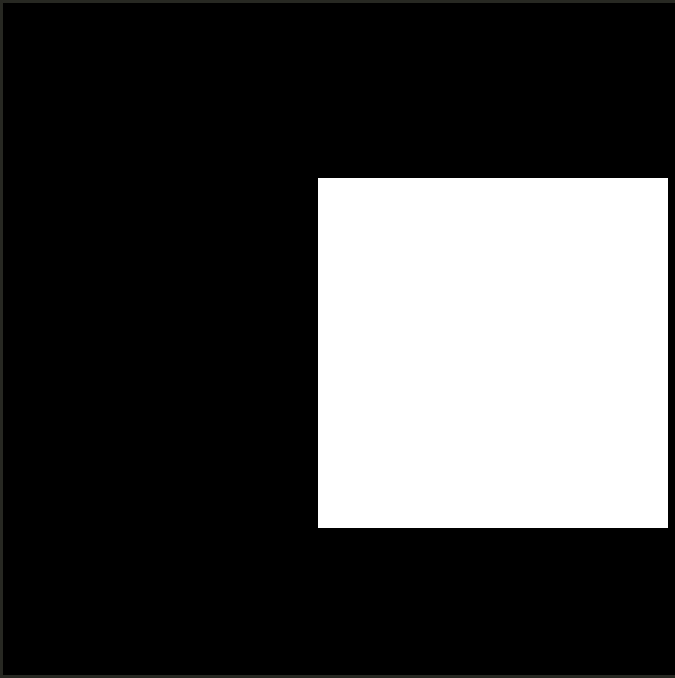}}
    \end{overpic}}
    \caption*{(b) A \textcolor{fgorange}{ship} docked at a port. }
    \end{subfigure}
    \caption{\textbf{Text to Image synthesis:} We augment Stable-Diffusion v2.1 with \methodwithspace to produce images with spatial control including the size and location of the objects. Inset images are the masks passed to the model. Best viewed when zoomed in.} %
    \label{fig:text2image}
\end{figure}

While \methodwithspace was designed for video synthesis, it can be easily modified and work for the task of Text-to-Image synthesis. Figure~\ref{fig:text2image} shows the versatility of our method. We generate images using Stable-Diffusion v2.1 \cite{Rombach_2022_CVPR} and gained spatial control through \method. We observe that for the same prompt and initialization seed, \methodwithspace is able to control the location of the subject making the generation process interactive. Please refer to appendix for more results.

\section{Implementation details}
\paragraph{ModelScope} - We generate videos of 256x256 resolution, and 16 frames. We fix the fixed step $t$ to be 2 for all generations for ssv2-ST and 4 for IMC generation, and diffusion steps to be 40. For numbers on IMC, we generate 24 frames. In the quality evaluation experiments Table~\ref{tab:qa}, we re-evaluated ModeScope performance on our selected set of prompts from the MSR-VTT dataset. \methodwithspace generation results are for videos generated with fixed step $t$ equal to 2 of 40 steps. 
\paragraph{ZeroScope}- We generate videos of 320 x 576 resolution, and 24 frames. We fix the fixed step $t$  to be 2 for all generations for ssv2-ST and 4 for IMC generation, and diffusion steps to be 40. 

\section{Ablation studies}

\subsection{Sensitivity to $t$}
\begin{figure*}[t!]
    \centering
        \begin{subfigure}[b]{0.49\linewidth} %
        \includegraphics[width=\linewidth]{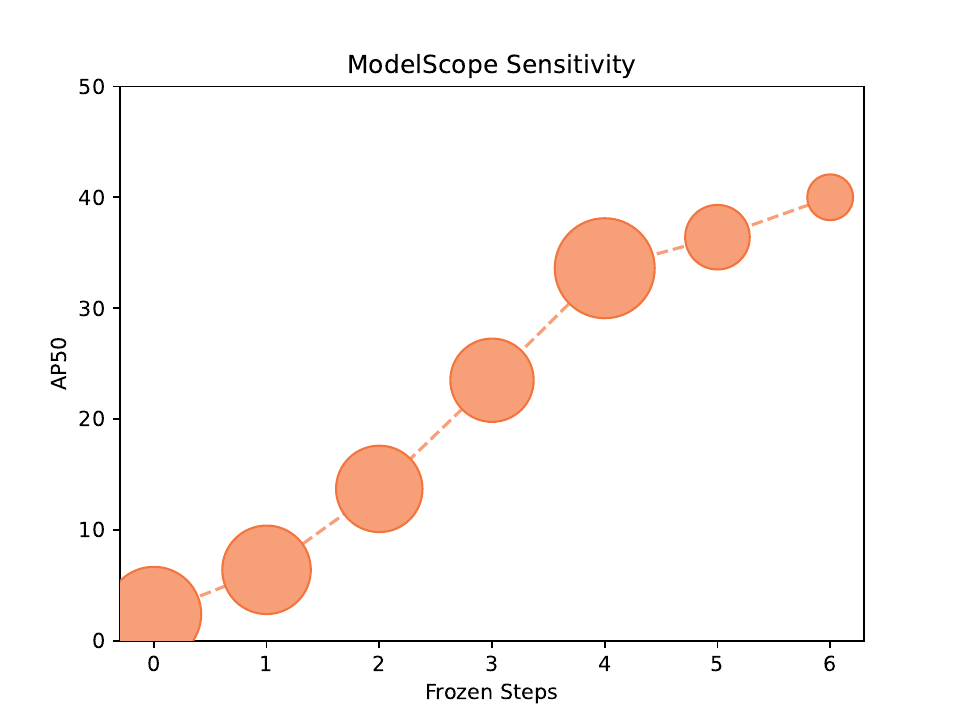}
    \end{subfigure}
    \hfill %
    \begin{subfigure}[b]{0.49\linewidth} %
        \includegraphics[width=\linewidth]{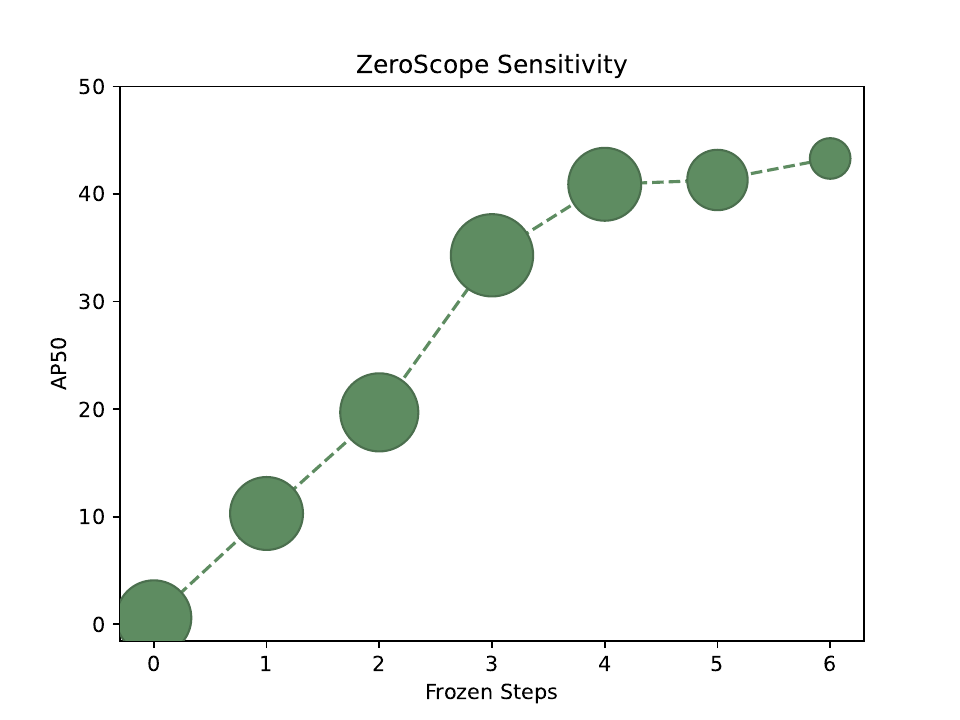} %
    \end{subfigure}

    \caption{\textbf{Sensitivity to frozen steps:} We plot AP50 against number of frozen steps $t$ for ModelScope and ZeroScope. The radius of the marker is proportional to the coverage. We find that increasing $t$ increases AP50 at the risk of losing coverage, i.e. degrading quality.}
    \label{fig:sensitivity_analysis}
\end{figure*}
In fig~\ref{fig:sensitivity_analysis}, we present results on varying the $t$ parameter for generation on the IMC dataset. As $t$ increases, AP50 increases, but coverage decreases. We hypothesize that this is because of selective attention masking for longer steps affecting the diffusion process. As the base models, were not trained with selective attention the model output quality degrades with larger number of steps.

\subsection{More results on masking}
We supplement the results of Fig~\ref{fig:ablations} with a full results table in Tab~\ref{tab:ablations}. As we can see from AP50, Cross Attn mask is responsible for the majority of the control as it determines the location of the foreground object in the canvas. While Self Attn mask is responsible for maintaining generated video quality as observed by Coverage scores.
\textbf{\begin{table}[]
\centering
\caption{\textbf{Ablation study on \methodwithspace}: We evaluate ModelScope without various attention masks on our user defined dataset. We find that each component of our method impacts the performance significantly} %
\label{tab:ablations} %
\resizebox{\columnwidth}{!}{%
\setlength{\tabcolsep}{4pt}
\centering
\begin{small}
\begin{tabular}{lcccc}
\toprule
Model           & mIoU \% ($\uparrow$) & Coverage \% ($\uparrow$)  & CD ($\downarrow$) & AP50 \% ($\uparrow$) \\
\midrule %
ModelScope+\methodwithspace        & 36.1 & 96.6     & 0.13 & 33.3  \\
-w/o Cross Attn Mask    & 14.2 & 93.3     & 0.27 & 5.7   \\
-w/o Self Attn Mask    & 19.5 & 87.7     & 0.30 & 16.5  \\
-w/o Temp Attn Mask & 19.7 & 96.6     & 0.25 & 9.9 \\
\midrule
ZeroScope+\methodwithspace        & 36.3 & 96.3 & 0.12 & 33.8  \\
-w/o Cross Attn Mask    & 13.3 & 78 & 0.23 & 6.7   \\
-w/o Self Attn Mask    & 25.6 & 83 & 0.21 & 28.7  \\
-w/o Temp Attn Mask & 25.1 & 91 & 0.18 & 18.6  \\
\bottomrule
\end{tabular}
\end{small}
}
\end{table}}

\section{Effect of subjects on video generation}

\begin{figure*}[t!]
    \centering
        \begin{subfigure}[b]{0.8\linewidth} %
        \includegraphics[width=\linewidth]{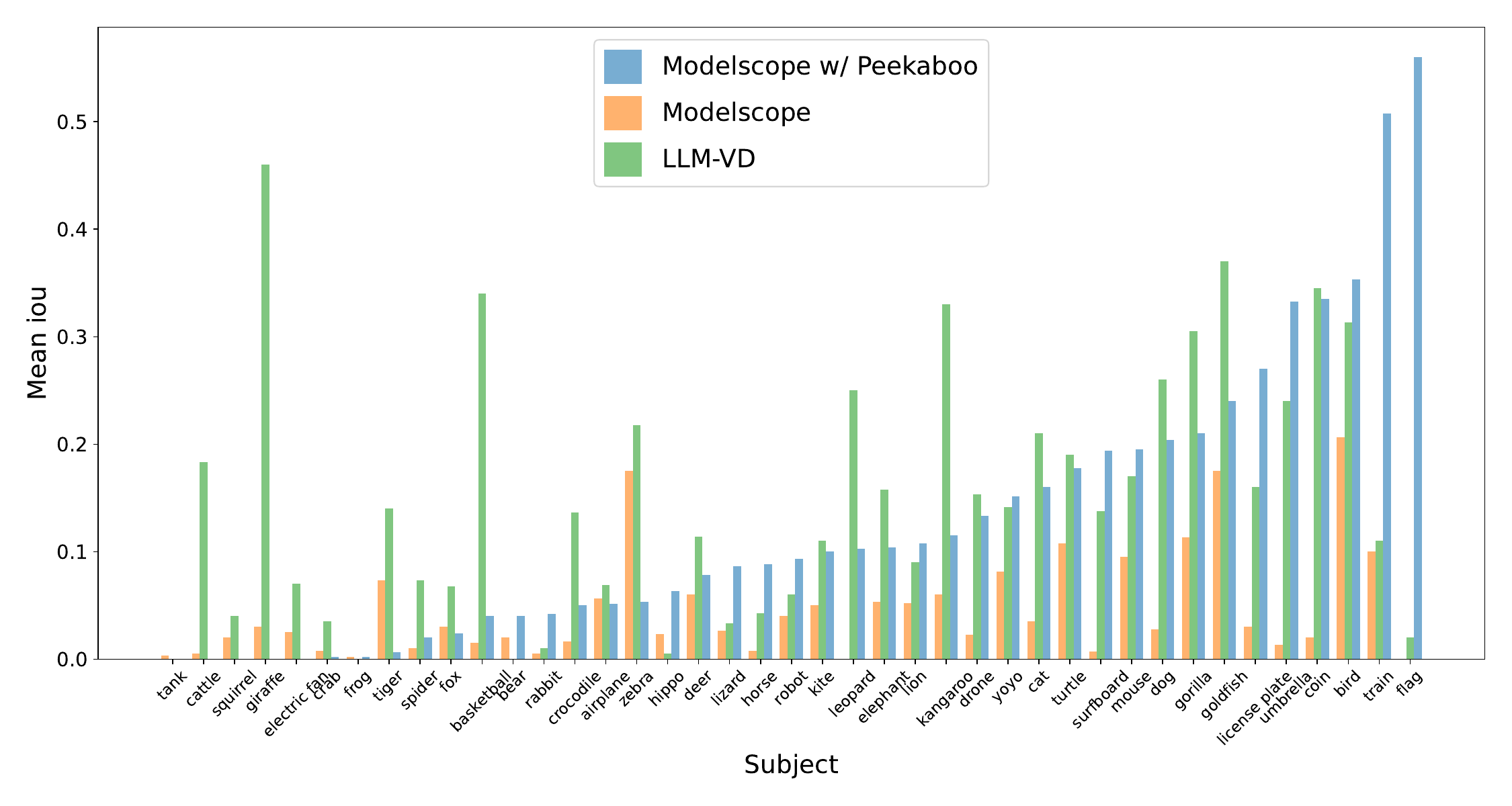}
    \end{subfigure}
    \caption{\textbf{Effect of subjects on video generation quality:} We plot mIoU on LaSOT against different subjects present in the dataset. We observe that \methodwithspace either retains or improves generation quality for majority of subjects. }
    \label{fig:subvsgeneration}
\end{figure*}

Evaluating quality across foreground subject depends heavily on the base model's training data. However, as we show in Fig~\ref{fig:miracle}, using masked diffusion in \methodwithspace can overcome model failures to some extent. We further evaluate the effect of subjects on controlled video generation by \methodwithspace against LaSOT dataset. Fig~\ref{fig:subvsgeneration} shows that \methodwithspace overcomes the base model's training bias and improves the control in video generation. Interestingly, \methodwithspace performance is better for some subject classes than other, perhaps it is a factor of disentanglement of CLIP embeddings used in cross-attention. Studying the cause of this imbalance warrants further research and we encourage the community to do so.

\section{More videos}
\begin{figure*}[t!]
    \centering
        \begin{subfigure}[b]{\linewidth} %
        \includegraphics[width=\linewidth]{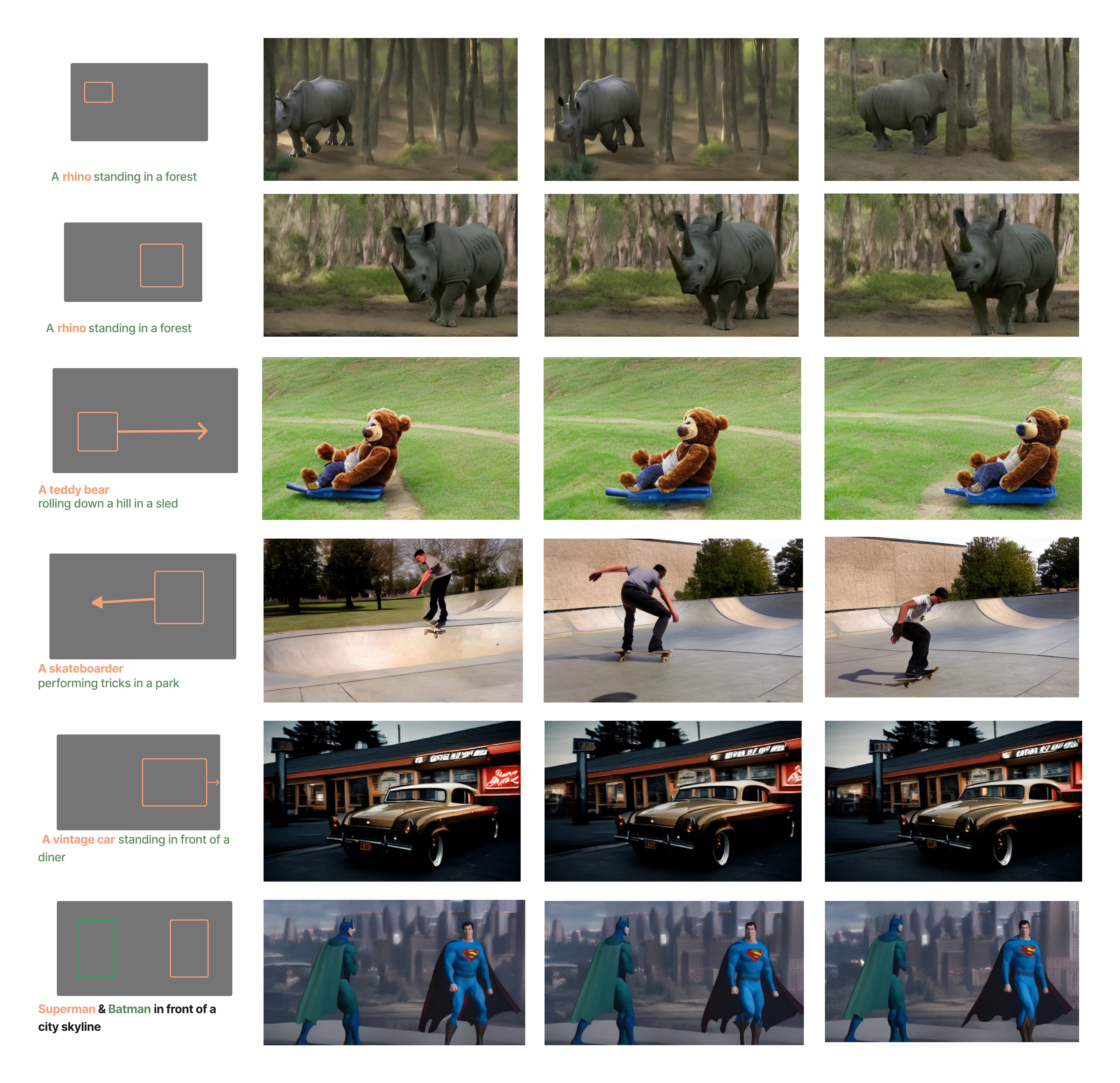}
    \end{subfigure}
    \caption{\textbf{More interactive video generation results}  }
    \label{fig:supp_videos}
\end{figure*}

We have uploaded the videos of the results presented in the main paper with our supplementary material and are available on the project webpage. Please use VLC media player for viewing the videos. We also append more video results in the Fig~\ref{fig:supp_videos} for the reader.
\section{Dataset Curation and filtering}

\subsection{IMC}
In this section, we provide finer details of generating IMC dataset.
\subsubsection{Prompts}
We provide the list of prompts used in IMC dataset.
\begin{itemize}

\item A woodpecker climbing up a tree trunk.
\item A squirrel descending a tree after gathering nuts.
\item A bird diving towards the water to catch fish.
\item A frog leaping up to catch a fly.
\item A parrot flying upwards towards the treetops.
\item A squirrel jumping from one tree to another.
\item A rabbit burrowing downwards into its warren.
\item A satellite orbiting Earth in outer space.
\item A skateboarder performing tricks at a skate park.
\item A leaf falling gently from a tree. 
\item A paper plane gliding in the air.
\item A bear climbing down a tree after spotting a threat.
\item A duck diving underwater in search of food.
\item A kangaroo hopping down a gentle slope.
\item An owl swooping down on its prey during the night.
\item A hot air balloon drifting across a clear sky.
\item A red double-decker bus moving through London streets.
\item A jet plane flying high in the sky.
\item A helicopter hovering above a cityscape.
\item A roller coaster looping in an amusement park.
\item A streetcar trundling down tracks in a historic district.
\item A rocket launching into space from a launchpad.
\item A deer standing in a snowy field.
\item A horse grazing in a meadow.
\item A fox sitting in a forest clearing.
\item A swan floating gracefully on a lake.
\item A panda munching bamboo in a bamboo forest.
\item A penguin standing on an iceberg.
\item A lion lying in the savanna grass.
\item An owl perched silently in a tree at night.
\item A dolphin just breaking the ocean surface.
\item A camel resting in a desert landscape.
\item A kangaroo standing in the Australian outback.
\item A colorful hot air balloon tethered to the ground.

\end{itemize}

\subsubsection{Generating the bounding Boxes}

Given the set of prompts, we annotate the main subject in the prompt. Further, the prompts are classified as stationary/moving, along with the object's aspect ratio as square, vertical rectangle, or horizontal rectangle. Specifically, the aspect ratio values are $1:1$, $4:3$, $3:4$ respectively. For prompts with movement, we also classify movement into up/down, left/right or zig-zag.

Three sets of bounding boxes are generated for each prompt. The starting co-ordinate of the bounding box is chosen randomly from 9 centroids of a 3x3 grid that the canvas is divided into. The speed is randomly chosen from 5-20 for moving prompts. The movement direction is randomly flipped as well. The bonding box size is chosen as 0.25 or 0.35 of the canvas size.  We then generate a bounding box for each frame according to the random parameters, adding a small jitter for each pixel is well. For moving prompts, the starting location is one of 6 centroids, omitting the centroids which align with the direction of motion. We will release the code for generating this dataset as well.

\subsection{ssv2-ST}
\paragraph{Filtering} -  We use Something-Something v2 dataset \cite{goyal2017something, mahdisoltani2018ssv2} to obtain the generation prompts and ground truth masks from real action videos. We filter out a set of 295 prompts. The details for this filtering are in the appendix. We then use an off-the-shelf \textit{OWL-ViT-large} open-vocabulary object detector \cite{owlvit} to obtain the bounding box annotations of the object in the videos. This set represents bounding box and prompt pairs of real-world videos, serving as a test bed for both the quality and control of methods for generating realistic videos with spatio-temporal control. 
We filter out the prompts such that they contain a single foreground object and obtain the bounding boxes or masks for the videos. We also further filter out videos with 0 bounding boxes. 

\paragraph{Post-processing bounding boxes} - We downsample videos in SSv2 to 5fps and 224x224 resolution. For each video, we consider the first 24 frames for computing bounding boxes. We use OwL-ViT/B16 for getting the bounding boxes of the first 24 frames. Due to frame jittering and low resolution, we observe that obtained masks were not consistently calculated for each frame. Hence, we interpolated the masks between two successive frames. Our final test set contains 295 prompts and masks pairs. We pass the first 16 of these boxes to ModelScope, and all 24 of them to ZeroScope

\section{Limitations}
\begin{figure*}[t!]
   \centering
    \hspace{0.1mm}
   \begin{subfigure}{\textwidth}
       \centering
        \includegraphics[width=\linewidth]{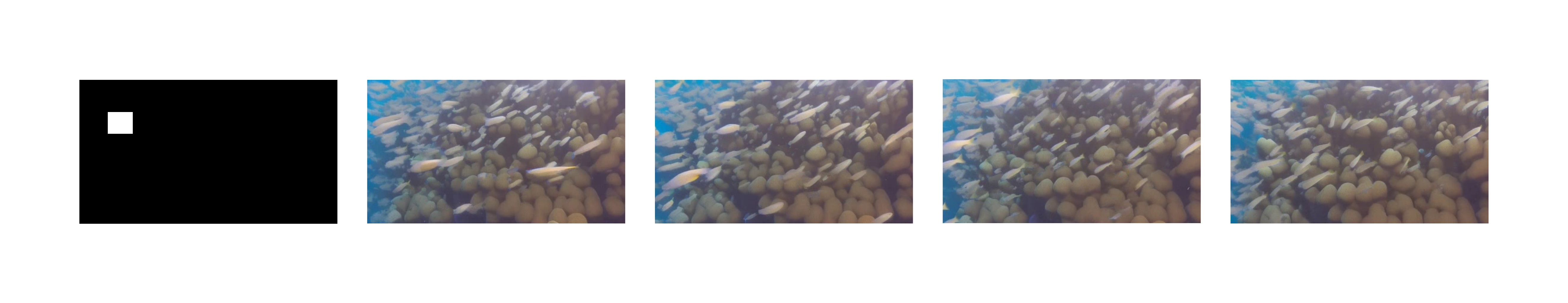}
       \caption{A \textcolor{fgorange}{school of fish} \color{black} in the ocean. }
   \end{subfigure}
   \vspace{-1em} %
   \begin{subfigure}{\textwidth}
       \centering
        \includegraphics[width=\linewidth]{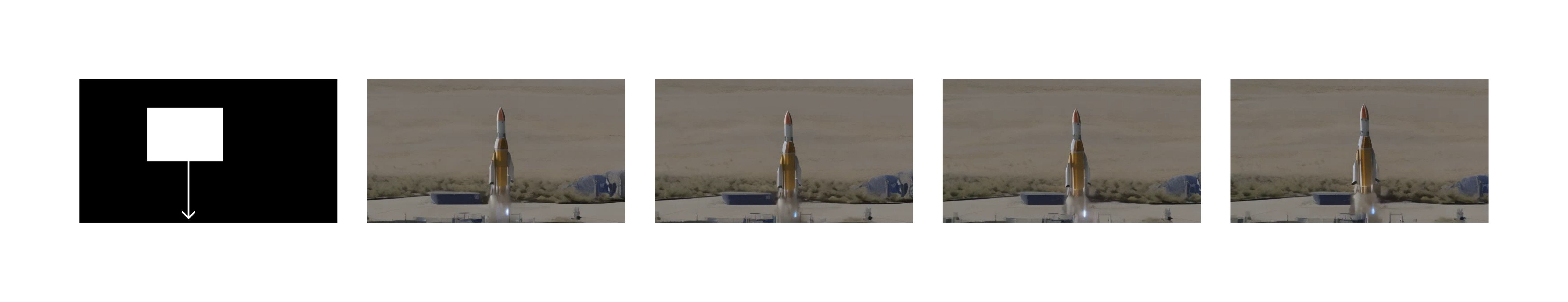}
       \caption{A \textcolor{fgorange}{rocket} \color{black} launching into space. }
   \end{subfigure}
   \vspace{-0.5em} %
   \begin{subfigure}{\textwidth}
       \centering
        \includegraphics[width=\linewidth]{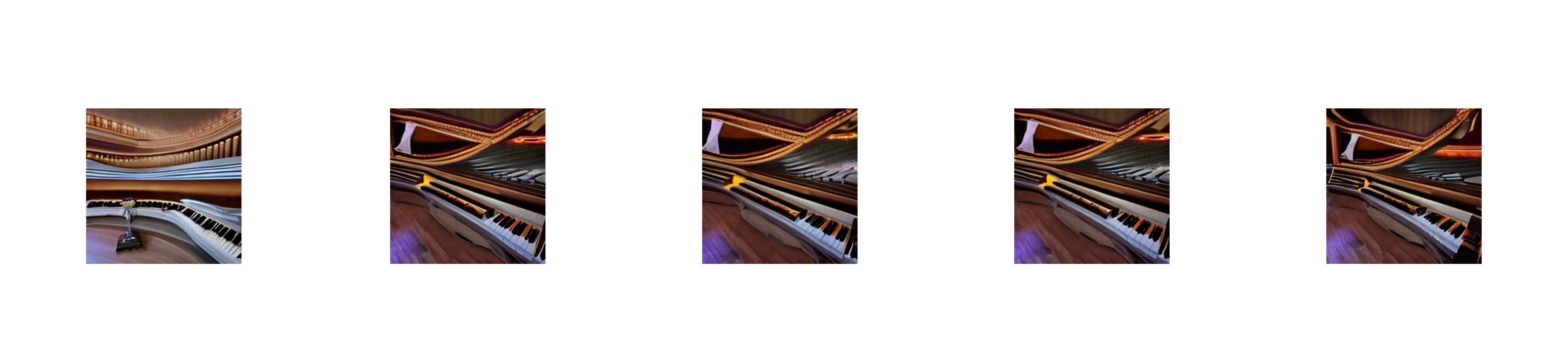}
       \caption{A \textcolor{fgorange}{grand piano} \color{black} in a hall. }
   \end{subfigure}
    
    \caption{\textbf{Our Failure modes}: Top row shows a failure mode because the mask is too small for the subject. Middle row shows a failure model where the object does not move much, since the direction of motion of the mask contradicts that of the text. Bottom row shows a case where the model inherits a bad generation of the base model. }
    \label{fig:failure}
\end{figure*}
In Fig~\ref{fig:failure}, we depict three typical failure modes of our method. These usually happen because there is a mismatch between the prior and the input mask, \textit{ie.}, the bounding boxes should be of sensible size that align with the training data of the model. Further, the generation usually fails for cases where the base model is bad at the target prompt. Moreover, the movement introduced through interactive control should align with the input text prompt.

\section{Societal Impact}
This is a work on controllable video generation and not video generation itself. It is possible that the base model itself reflects some societal biases of the training set which will be propagated with the work. It also inherits the potential for misuse that other such video generation works have.

\end{document}